# Machine Learning Emulation of 3D Cloud Radiative Effects

David Meyer[1,2] (ORCID: 0000-0002-7071-7547)

Robin J. Hogan[3,1] (ORCID: 0000-0002-3180-5157)

Peter D. Dueben[3] (ORCID: 0000-0002-4610-3326)

Shannon L. Mason[3,1] (ORCID: 0000-0002-9699-8850)

[1]Department of Meteorology, University of Reading, Reading, UK

[2]Department of Civil and Environmental Engineering, Imperial College London, London, UK

[3]European Centre for Medium-Range Weather Forecasts, Reading, UK

Correspondence to David Meyer (d.meyer@pgr.reading.ac.uk)

**Key Points:**

- Emulators are used to add 3D cloud effects to the fast 1D radiation solver Tripleclouds cheaply
- Emulators are trained on 3D cloud effects from SPARTACUS, which is five times slower than Tripleclouds
- For a 1 % slowdown in Tripleclouds' runtime, 3D fluxes are emulated with an average error of about 20 %–30 %

**Abstract**

The treatment of cloud structure in numerical weather and climate models is often greatly simplified to make them computationally affordable. Here we propose to correct the European Centre for Medium-Range Weather Forecasts 1D radiation scheme *ecRad* for 3D cloud effects using computationally cheap neural networks. 3D cloud effects are learned as the difference between ecRad's fast 1D *Tripleclouds* solver that neglects them and its 3D *SPARTACUS* (SPeedy Algorithm for Radiative TrAnsfer through CloUd Sides) solver that includes them but is about five times more computationally expensive. With typical errors between 20 % and 30 % of the 3D signal, neural networks improve Tripleclouds' accuracy for about 1 % increase in runtime. Thus, rather than emulating the whole of SPARTACUS, we keep Tripleclouds unchanged for cloud-free parts of the atmosphere and 3D-correct it elsewhere. The focus on the comparably small 3D correction instead of the entire signal allows us to improve predictions significantly if we assume a similar signal-to-noise ratio for both.

**Plain Language Summary**

Solar and terrestrial radiation is the primary driver of the Earth's weather and their detailed representation is essential for improving weather predictions and climate projections. Several aspects, however, such as the flow of radiation through the side of clouds and other three-dimensional effects are often too costly to compute routinely. In this paper we describe how machine learning can help account for these effects cheaply.

1. Introduction

Solar (hereafter, shortwave) and thermal-infrared (hereafter, longwave) radiation are the primary drivers of atmospheric weather systems via their role in creating the equator-to-pole temperature gradient, while their interaction with greenhouse gases drives anthropogenic climate change. As such, their detailed representation in radiation schemes is essential for both weather and climate (e.g., Stephens 1984).

Historically viewed as one of the slowest components of atmospheric models, radiation schemes have been among the prime candidates for acceleration via machine learning (ML). Cheruy et al. (1996) are one of the first to develop a neural network (NN) emulator of longwave radiation, reporting accurate results for speedups of 3 and 3 000 times the reference broadband and narrowband models, respectively. Chevallier et al. (1998, 2000) extend Cheruy et al. (1996)'s work to the European Centre for Medium-Range Weather Forecasts' (ECMWF) 50-level longwave and shortwave radiation scheme, reporting promising results for a sixfold reduction in computational costs. Krasnopolsky et al. (2005) develop a longwave NN emulator in the NCAR's (National Center for Atmospheric Research) Community Atmosphere Model, reporting predicted heating rate root-mean-square errors between 0.26 and 0.33 K d$^{-1}$, for a runtime reduction between 35 and 80 times the original schemes, while Krasnopolsky et al. (2008)'s emulation of shortwave radiation report successful multi-decadal simulations with offline scheme speedups of 150 and 20 times the original scheme, and errors of 0.34 and 0.19 K d$^{-1}$ for longwave and shortwave heating rates, respectively. More recently, Pal et al. (2019)'s emulation of shortwave and longwave radiation in the Super-Parameterized Energy Exascale Earth System Model report a speedup of about one order of magnitude while retaining 90 %–95 % of the original scheme's accuracy.

Although these findings are encouraging, an assessment of Chevallier et al. (1998, 2000)'s emulators by Morcrette et al. (2008) report degraded accuracy and performance when the number of levels increases above the original 50 levels. Ukkonen et al. (2020) note that differences in radiative fluxes may sometimes be larger than the internal variability of the original scheme or with regional errors in annual-mean surface net fluxes of 20 W m$^{-2}$. Furthermore, although Roh and Song (2020) report average root-mean-square errors of 1.0 and 0.49 K d$^{-1}$ for the longwave and shortwave heating rates, respectively, and 1.6 and 14 W m$^{-2}$ for the longwave and shortwave fluxes, respectively, large deviations of about 20 W m$^{-2}$ occur. Indeed, comparing these results is challenging as studies report their results using specific datasets and summarize them with different statistical metrics.

An important point to note when seeking applications of ML in radiative transfer is that radiation schemes are no longer a particularly slow component of atmospheric models. For example, in the ECMWF Integrated Forecast System (IFS), Hogan et al. (2017) report that the fractional time in the highest operational resolution model dropped from 19% in 2007 to 5% in 2017. Nevertheless, several simplifications are still made. For example: (a) operational radiation schemes cannot afford to represent the $O(10^5)$ spectral lines explicitly and typically approximate the spectral variation of gas absorption by $O(10^2)$ quasi-monochromatic radiative transfer calculations (Hogan and Matricardi 2020); (b) the accuracy of radiative forcing calculations due to changes in greenhouse gases in many schemes is questionable, with Soden et al. (2018) reporting that the spread of radiative forcing estimates due to increased $CO_2$ is about 35 % of the mean; and (c) the ways in which radiation interacts with cloud structure in radiation schemes is generally quite crude. Specifically, all radiation schemes used routinely in weather and climate models are "1D"; that is, they neglect the full 3D interaction of radiation with clouds. They, therefore, ignore the interception of direct sunlight by cloud sides, the trapping of sunlight beneath clouds, and the emission of thermal radiation from the sides of clouds (e.g., Hogan & Shonk, 2013; Varnai & Davies, 1999). For cumulus clouds, these missing processes, here defined as 3D cloud effects, can change the magnitude of instantaneous cloud radiative effects (i.e., the difference between fluxes in the presence of clouds and in the equivalent clear-sky conditions) by 30 % in the longwave (e.g., Heidinger and Cox 1996) and by up to 60 % in the shortwave, depending on sun angle (e.g., Pincus et al. 2005). Although the method of Jakub and Mayer (2015) may be more appropriate for (sub)kilometer-resolution where radiative exchanges between atmospheric columns become important, to date, the fastest method we are aware of to represent 3D interactions of radiation with clouds within a model column suitable for large-scale models with horizontal resolution no finer than 5–10 km is the Speedy Algorithm for Radiative Transfer through Cloud Sides (SPARTACUS; Hogan et al., 2016). Despite this, SPARTACUS is approximately five times slower than the radiation scheme currently used at the ECMWF (Hogan & Bozzo, 2018)—far too slow to be considered for operational use.

As an alternative method to the emulation of an entire radiation scheme, Ukkonen et al. (2020) and Veerman et al. (2021) show a different approach whereby specific parts of a scheme, such as the gas optics in the RTE-RRTMGP (Radiative Transfer for Energetics and Rapid and accurate Radiative Transfer Model for General circulation models applications – Parallel; Pincus et al., 2019) framework, are emulated, retaining the original radiative transfer solver. Similar to the results by Veerman et al. (2021), Ukkonen et al. (2020) report heating rates and top-of-atmosphere longwave and shortwave radiative forcing root-mean-square errors relative to benchmark line-by-line radiation calculations typically below 0.1 K d$^{-1}$ and 0.5 W m$^{-2}$, respectively, relative to benchmark line-by-line radiation calculations (with smaller errors relative to RTE-RRTMGP), and speedups for clear-sky longwave and shortwave fluxes of 3.5 and 1.8 times the original scheme, respectively.

In the same spirit, here we investigate how to improve the representation of 3D cloud effects with a hybrid physical-ML method. Rather than replacing the entire radiation scheme, we run the existing 1D radiation scheme in parallel with an emulator trained on the difference between SPARTACUS and the 1D scheme. As the computational cost of the emulator is expected to be a fraction of that of SPARTACUS, we hope to achieve a similar accuracy for a fraction of the cost. Moreover, as heating rates are susceptible to vertical changes in fluxes, by only correcting profiles within the troposphere, we expect this approach to be more tolerant to errors in heating rates for higher parts of the atmosphere where low values of atmospheric pressure exacerbate comparatively small errors in predicted fluxes.

The following sections describe the general method, with specific details about reference models and data (Section 2.1) used to develop and train the ML emulators (Section 2.2). We follow with a qualitative (Section 3.1) and quantitative (Section 3.2) evaluation of the results, as well as a runtime performance analysis of the emulators (Section 3.3), before concluding with a summary and prospects for future work (Section 4).

## 2. Methods

### 2.1 Reference Model and Data

Reference simulations use the open-source atmospheric radiative transfer software ecRad (Hogan & Bozzo, 2018) version 1.3.0 (ECMWF, 2020). ecRad computes profiles of up- and downwelling, long- and shortwave radiative fluxes (with downwelling shortwave having both total and direct components) from zero- or one-dimensional (i.e., profiles) inputs of meteorological variables such as dry-bulb air temperature, cloud fraction, mixing ratios of water vapor, liquid water, ice cloud, snow, and trace gases. 3D cloud effects are computed as the difference between ecRad's 3D solver SPARTACUS and ecRad's 1D solver Tripleclouds (Shonk & Hogan, 2008). Although deterministic forecasts in the ECMWF Integrated Forecast System (IFS) use ecRad's 1D solver McICA (Monte Carlo Independent Column Approximation; Pincus et al. 2003), Tripleclouds is used here as (a) its flux predictions are noise-free, and (b) its underlying assumptions in cloud structure and overlap are the same as in SPARTACUS. Here, ecRad is forced with inputs from the EUMETSAT Numerical Weather Prediction Satellite Application Facility (NWP-SAF) data set (Eresmaa & McNally, 2014). This data set contains 25 000 atmospheric profiles representative of yearly, global, present-day atmospheric conditions on 137 atmospheric levels (surface to 0.01 hPa) from ECMWF operational forecasts between 2013 and 2014. Profiles of aerosol mixing ratio and greenhouse gas concentration are from the climatology of Bozzo et al. (2020) as a function of longitude, latitude, and month for the former, and latitude and month, for the latter. The prescribed horizontal cloud scale in SPARTACUS uses the parameterization of Fielding et al. (2020).

### 2.2 Neural Network Emulator

Two separate NNs to emulate short- and longwave 3D cloud effects are developed using the multilayer perceptron (MLP)—a standard form of NN with inputs traveling via one or more hidden layers towards the outputs (Bishop, 2006)—following poor results from a preliminary investigation using linear regression (not reported). Both NNs are implemented in Python with TensorFlow (Abadi et al., 2015) version 2.4.1.

To capture the interaction of radiation with clouds, we compute the cloud optical depth $\tau_c$ in the large particle limit where geometric optics is applicable, albeit ignoring small spectral dependences, as $\frac{3}{2}\frac{\Delta p}{g}\left(\frac{q_l}{\rho_l r_l} + \frac{q_i}{\rho_i r_i}\right)$, with $\Delta p$ the difference in atmospheric pressure between two atmospheric layers, $g$ the gravitational acceleration constant (9.81 m s$^{-2}$), $q_l$ and $q_i$ the liquid and ice mass mixing ratios, $\rho_l$ and $\rho_i$ the densities of liquid water and ice, and $r_l$ and $r_i$ the liquid and ice effective radii. Gas and aerosol properties, known to affect 3D effects minimally, are ignored. Heating rates $dT/dt$ are computed from the *net* (downwelling *minus* upwelling) flux for an atmospheric layer $i$ as $-\frac{g}{c_p}\frac{F^n_{i+1/2}-F^n_{i-1/2}}{p_{i-1/2}-p_{i+1/2}}$, with $p_{i\pm1/2}$ and $F^n_{i\pm1/2}$ the atmospheric pressure and net flux, respectively, at the layer interface $i + 1/2$ and $i - 1/2$ (counting down from the top of the atmosphere), and $c_p$ the specific heat of dry air (1004 J kg$^{-1}$ K$^{-1}$). As heating rates are proportional to the vertical derivative of the net flux, noise in predicted fluxes can amplify the errors in computed heating rates. Although training the NNs using fluxes and heating rates can partially mitigate this issue, predictions can no longer conserve energy. To avoid this issue, here we instead predict the 3D *scalar* (downwelling *plus* upwelling) flux and corresponding heating rates, as well as the direct downwelling shortwave flux, and postprocess the outputs in a separate step (Appendix A) to obtain energy-consistent downwelling and upwelling fluxes and heating rates. As we aim to predict the 3D cloud effects, only levels between the surface and 50 hPa (i.e., assuming no clouds above the troposphere) are used. The full profiles, spanning 137 levels, are recovered by setting values between 50 and 0 hPa to zero for the downwelling component, and extending the last predicted value at 50 hPa to all levels between 50 and 0 hPa for the upwelling component.

Train, validation, and test datasets contain a random 60 % (13 702), 20 % (4 568), 20 % (4 568) selection of NWP-SAF profiles as inputs, and corresponding ecRad computed 3D cloud effect profiles (SPARTACUS minus Tripleclouds; Section 2.1) as outputs. Before being fed to the NNs, profiles are reshaped to two-dimensional matrices with each profile as row (sample) and flattened level and quantity as column (feature). To determine the sensitivity to different choices of hyperparameters and input quantities, a grid search is conducted. In it, NNs are trained with NWP-SAF profiles of dry-bulb air temperature $T$, cloud fraction $f_c$, surface temperature $T_s$, surface albedo $\alpha$, cloud optical depth $\tau_c$, cosine of solar zenith angle $\mu_0$, specific humidity $q$, and vertical layer thickness $\Delta z$ as inputs (Table 1a), and corresponding ecRad-computed 3D cloud effect profiles of scalar fluxes, heating rates, and direct downwelling shortwave as outputs (Table 1b). All configurations use the Exponential Linear Unit activation function, Adam optimizer with mean squared error on all outputs, and 1 000 epoch-limit with early stopping patience set to 50 epochs. The surface emissivity $\varepsilon$ is not used as it is constant across profiles. Iterations are repeated 10 times to account for the stochasticity of the training algorithm. Hyperparameter choices are (a) input quantities: $\{\{f_c,\tau_c,T,T_s,\alpha,\mu\},\{f_c,\tau_c,T,T_s,\alpha,\mu,q\},\{f_c,\tau_c,T,T_s,\alpha,\mu,q,\Delta z\}\}$; (b) number of hidden layers: {1, 2, 3, 4, 5}; (c) hidden (neuron) size multipliers: {0.5, 1, 2}; (d) L1 and L2 regularization factors: {10$^{-6}$, 10$^{-5}$, 10$^{-4}$}. The number of neurons in hidden layers is computed by multiplying the number of inputs (182 for shortwave and 271 for longwave) by the hidden size multiplier. Results are visually inspected (Figure 1) and the simplest NN configuration (e.g., fewer neurons and input quantities) with the lowest mean absolute is chosen. For both longwave and shortwave components, this 'optimal' configuration is found to have three hidden layers, each with 217 and 182 neurons per hidden layer for longwave and shortwave, respectively, and L1 and L2 regularization factors set to 10$^{-5}$ (Figure 1). The most sensitive input quantities are: $f_c$, $T$, $T_s$, $\alpha$, $\tau_c$, and $\mu_0$ (Table 1a). $q$ and $\Delta z$ are not used as they do not improve predictions (Figure 1); This is reasonable as (a) the cloud layer optical depth, which is proportional to layer thickness (for same cloud water mixing ratio), is already an input variable to the NNs, and (b) any dependence on humidity is likely captured by the dry-bulb air temperature for cloudy parts of the atmosphere. An increase in either the network size or the number of layers does not improve the overall accuracy (Figure 1). Convergence is achieved after approximately 100 epochs.

To improve the results further, the two NNs with the above-determined configuration are trained with more data. For this, we use Synthia (Meyer & Nagler, 2021) version 0.3.0 (Meyer & Nagler, 2020) as outlined in Meyer, Nagler, et al. (2021) but only for independent inputs. Thus, we (a) generate nine synthetic copies of the surface albedo $\alpha$ and cosine of the solar zenith angle $\mu_0$, (b) randomly re-assign them to plain copies of NWP-SAF train-fraction profiles, and (c) collate them together to the original data set to form a total of 137 020 profiles for training (i.e., the original 13 702 profiles and 123 318 modified profiles). These augmented profiles are then used in ecRad to generate corresponding training outputs, and both augmented inputs and outputs to train the NNs. To account for the variability in the results given by the NN's training algorithm, training (and inference) is run 20 times (10 times with and 10 times without data augmentation), varying random seeds between repeats. From this, the short- and longwave emulator with median mean absolute error are chosen. With this simple augmentation, the shortwave error is found to improve by about 18 %.

**Table 1.** Inputs and outputs used in the two NN emulators. Vector quantities are either at the interface between two model layers (half level; HL), or at the model layer (full level; FL). The superscript "$^L$" or "$^S$" denotes if the input is used in the longwave or shortwave NN. The scalar flux is defined as downwelling plus upwelling flux.

| Symbol | Name | Unit | Dimension |
|---|---|---|---|
| *(a) Inputs* | | | |
| $f_c$ | $^{L,S}$Cloud fraction | 1 | FL |
| $\tau_c$ | $^{L,S}$Cloud optical depth | 1 | FL |
| $T$ | $^L$Dry-bulb air temperature | K | FL |
| $T_s$ | $^L$Surface temperature | K | Scalar |
| $\alpha$ | $^S$Surface (shortwave) albedo | 1 | Scalar |
| $\mu_0$ | $^S$Cosine of solar zenith angle | 1 | Scalar |
| *(b) Outputs* | | | |
| $L^S$ | 3D effect on scalar longwave radiative flux density | W m$^{-2}$ | HL |
| $S^S$ | 3D effect on scalar shortwave radiative flux density | W m$^{-2}$ | HL |
| $S^\Downarrow$ | 3D effect on downwelling direct shortwave radiative flux density | W m$^{-2}$ | HL |
| $L^H$ | 3D effect on longwave heating rate | K s$^{-1}$ | FL |
| $S^H$ | 3D effect on shortwave heating rate | K s$^{-1}$ | FL |

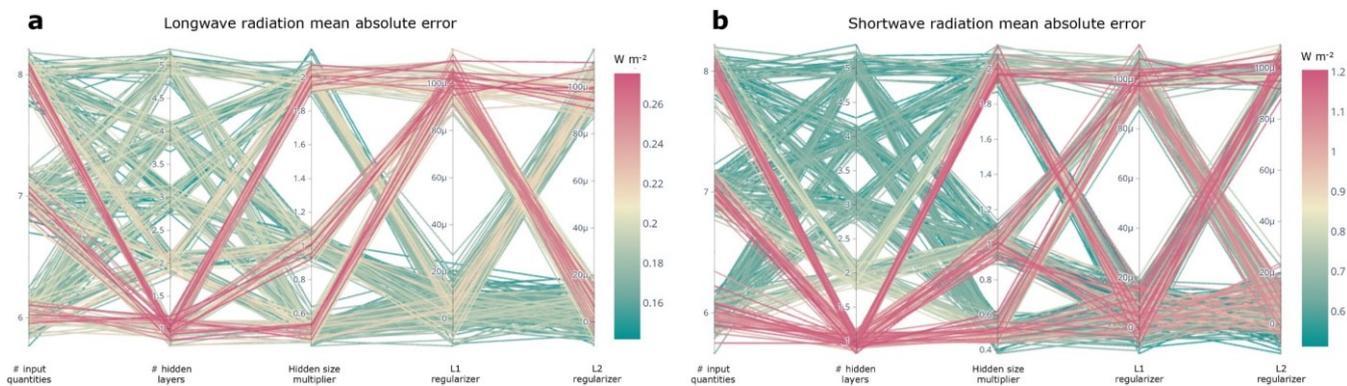

**Figure 1.** Mean absolute errors resulting from different hyperparameters configurations for the (**a**) longwave and (**b**) shortwave neural network. Each line represents a realization from a different hyperparameter configuration. Lines are shown slightly offset in the vertical axis for clarity. The search is conducted for configurations of (i) input quantities: $\{\{f_c, \tau_c, T, T_s, \alpha, \mu\}, \{f_c, \tau_c, T, T_s, \alpha, \mu, q\}, \{f_c, \tau_c, T, T_s, \alpha, \mu, q, \Delta z\}\}$ shown as 6, 7, and 8, respectively; (ii) hidden layer size: {1, 2, 3, 4, 5}; (iii) hidden (neuron) size multipliers: {0.5,1,2}; and (iv) L1 and L2 regularization factors: {10$^{-6}$, 10$^{-5}$, 10$^{-4}$}. Hidden size multipliers are multiplied by the number of inputs (182 for shortwave and 271 for longwave) to obtain the number of neurons in each hidden layer.

## 3. Results and Discussion

### 3.1 Qualitative Evaluation

First, a separate visual inspection is conducted using an atmospheric slice of ERA5 reanalysis data (Hersbach et al., 2020), extending from north to south poles at a longitude of 5 °E at 12:00 UTC (Coordinated Universal Time) on 11 July 2019. This includes the response of radiation to Saharan dust, marine stratocumulus, deep convection, and Arctic stratus. The surface albedo, cosine of the solar zenith angle, and cloud fraction are shown in Figure 2. Figure 3 shows the outputs from SPARTACUS (left), reference 3D cloud effects (3D signal; SPARTACUS minus Tripleclouds; middle), and NN predictions (right), respectively. The longwave effect of clouds (Figures 3a and 3d) is to warm the Earth system by reducing the upwelling radiation to space and increasing it towards the surface. When the 3D effects are simulated, clouds can not only interact with radiation through their base and top, but also through their sides. Thus, they further reduce the upwelling longwave radiation to space and further increase it towards the surface (Figures 3b and 3e). Figure 3n shows that the longwave heating rate signal of clouds is also amplified, increasing the magnitudes of cooling at cloud tops, and of warming at cloud bases (see Schafer et al., 2016 for further discussion). In the shortwave, the sign of the impact is dependent on solar zenith angle (Figure 3k): when the Sun is near its zenith, at the tropics, the 3D cloud effect acts to reduce the upwelling radiation reflected into space from cloud tops, but to increase it when near the horizon, over the Southern Ocean. These behaviors can be explained by the mechanisms of entrapment and side-illumination, respectively (Hogan et al., 2019). Although the vertical structure of heating rate is smoothed somewhat vertically in both the longwave (Figure 3o) and shortwave (Figure 3r), the sign and size predicted by the NN is captured for high and low clouds and for high and low sun angles. Figure 4 compares the 3D effects at top-of-atmosphere (TOA) upwelling fluxes and surface downwelling fluxes between the reference ecRad calculations and NN-predicted 3D cloud effects with generally good agreement across the range of latitudes.

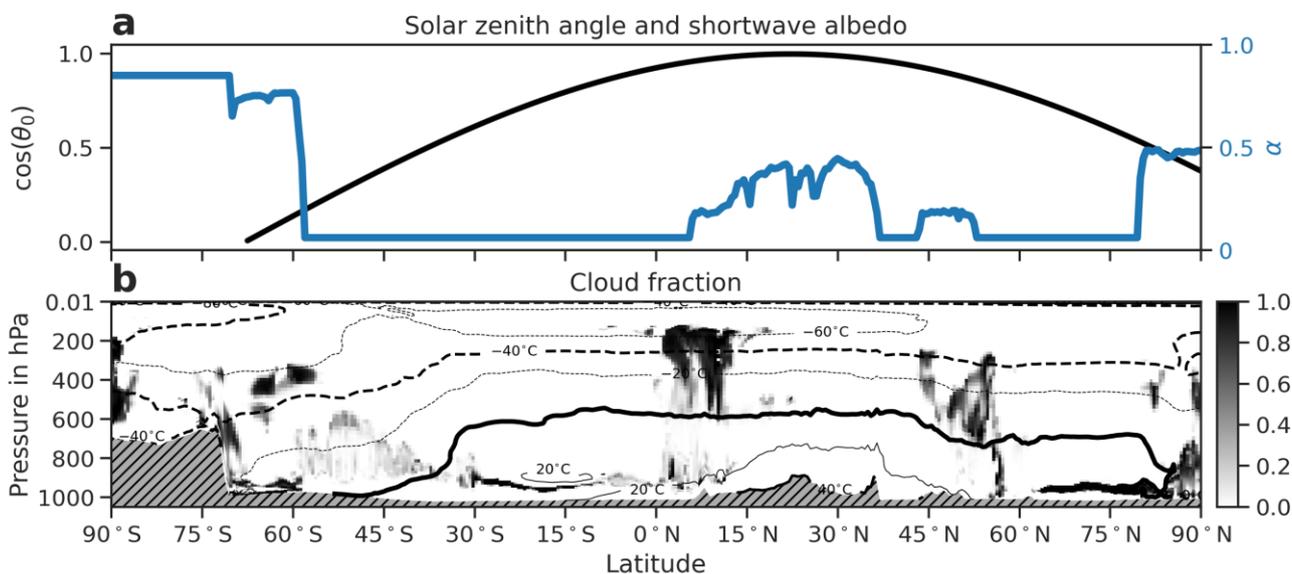

**Figure 2.** Typical zero- and one-dimensional SPARTACUS and Tripleclouds inputs: (**a**) surface albedo ($\alpha$; blue line), and cosine of the solar zenith angle $\cos(\theta_0)$; black line; scalar quantities), and (**b**) cloud fraction (vector quantity) at 5°E on 11 July 2019 12:00 UTC from ERA5 reanalysis data (Hersbach et al., 2020). Vector quantities consist of 137 vertical levels, here shown using atmospheric pressure as coordinate. The hatched area shows the topography. Temperature contours are shown using dashed and dotted lines.

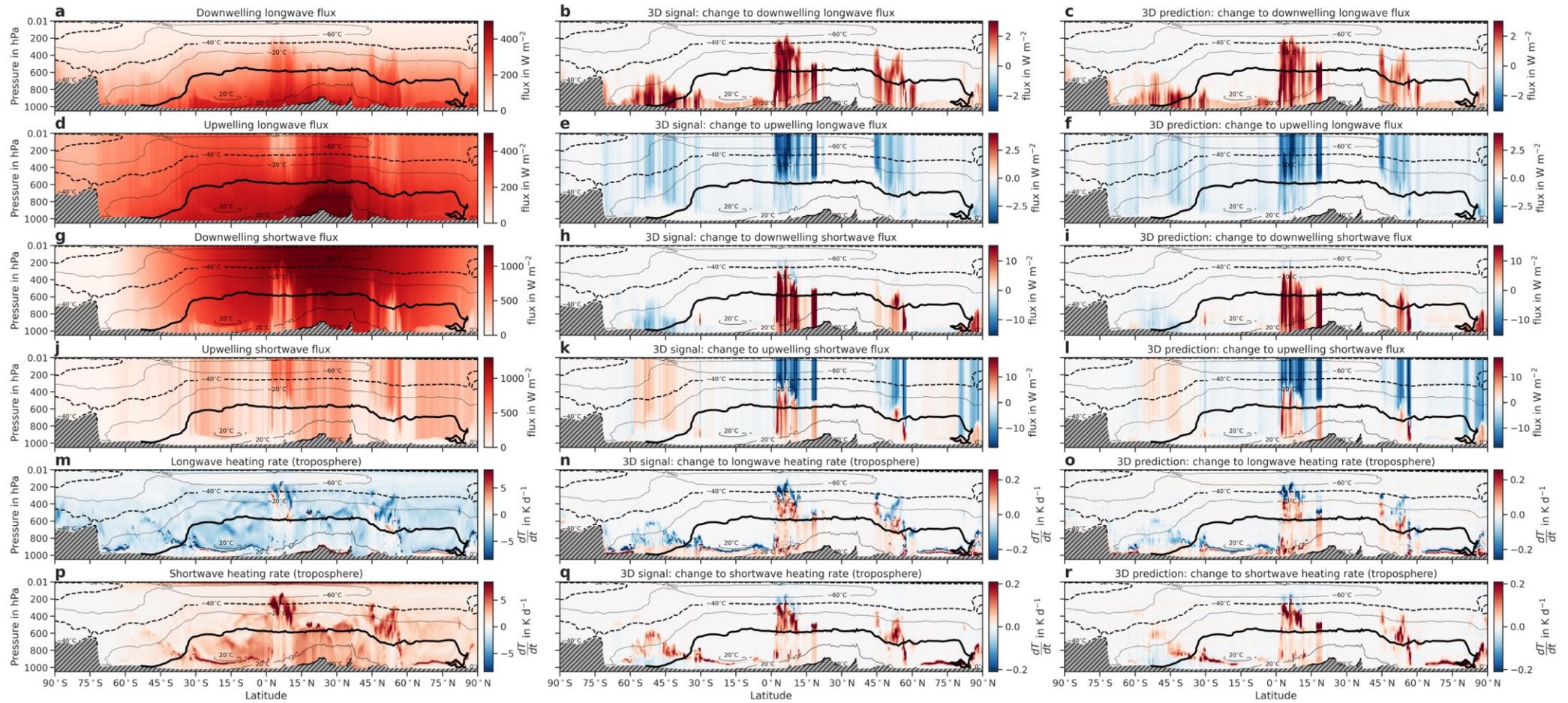

**Figure 3.** (**a-f**) Longwave and (**g-l**) shortwave fluxes, and (**m-o**) longwave and (**p-r**) shortwave heating rates, computed with pole-to-pole slice at 5°E on 11 July 2019 12:00 UTC from ERA5 reanalysis data (Hersbach et al., 2020), of (**a,d,g,j,m,p**) SPARTACUS, (**b,e,h,k,n,q**) 3D signal (SPARTACUS minus Triplecloud), and (**c,f,I,l,o,r**) 3D NN predictions. The slice includes the response of radiation to Saharan dust, marine stratocumulus, deep convection, and Arctic stratocumulus.

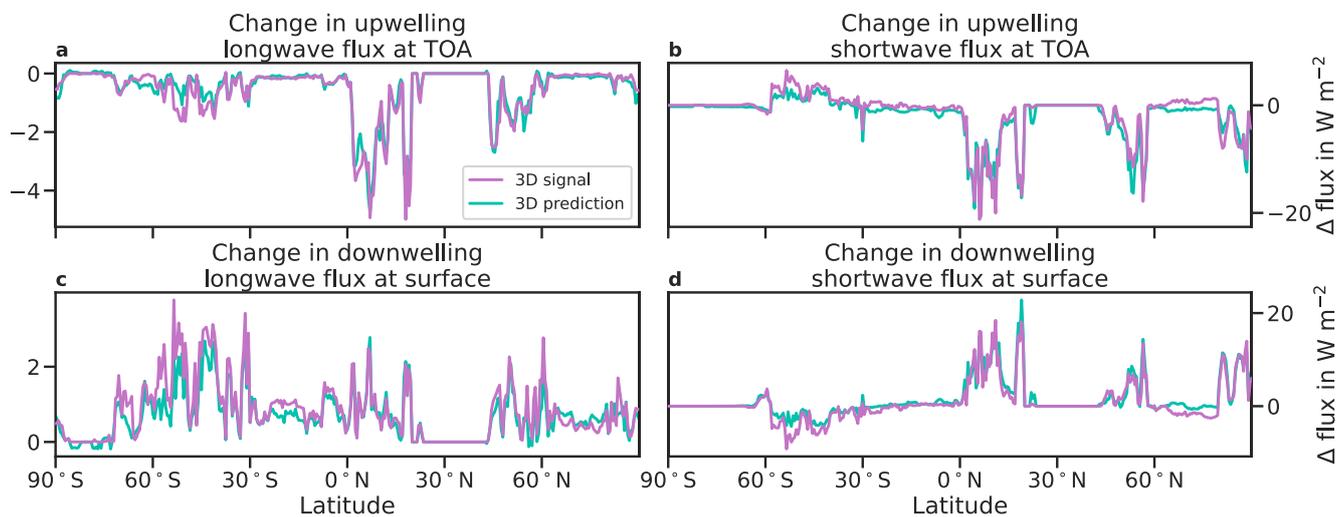

**Figure 4.** Comparison of 3D signal (SPARTACUS minus Tripleclouds; magenta) and 3D prediction (NN; cyan) for (**a-b**) top-of-atmosphere (TOA) upwelling flux and (**c-d**) surface downwelling flux using pole-to-pole slice at 5°E on 11 July 2019 12:00 UTC from ERA5 reanalysis data (Hersbach et al., 2020).

### 3.2 Quantitative Evaluation

Second, a quantitative evaluation is made by comparing NN-predicted 3D cloud effects to reference ecRad calculations (3D signal; SPARTACUS minus Tripleclouds) using the test fraction (Section 2.1). Evaluation metrics are computed using profiles of either 3D signal, 3D predictions, or error (i.e., prediction minus signal) separately for each level, or with no distinction in vertical levels (hereafter referred to as bulk), for a vector $y$ of $1 \ldots N$ samples for the mean ($\frac{1}{N}\sum_{i=1}^{N} y_i$) or mean absolute ($\frac{1}{N}\sum_{i=1}^{N} |y_i|$). Per-level statistics are shown in Figure 5 for fluxes, and in Figure 6 for heating rates. Bulk error statistics are summarized in Table 2. The first column in Figure 5 shows per-level means, and 50 % and 90 % quantiles of 3D signal and NN predictions. On average, the 3D signal is approximately 1 W m$^{-2}$ for the longwave (Figures 5a and 5d) and about 3 W m$^{-2}$ at the surface for the shortwave (Figure 5j). To put these numbers into context, the radiative forcing from doubling carbon dioxide concentrations from preindustrial levels is around 3.7 W m$^{-2}$ (Forster et al. 2007). Visually, NN predictions are close to the mean reference 3D signal (Figures 5a, 5d, 5g, 5j, and 5m). The 3D error for the mean (solid) and mean absolute (dashed) is shown in the second column of Figure 5. This reaches about 0.2 W m$^{-2}$ for the longwave (Figures 5b and 5e) and about 0.6 W m$^{-2}$ for the shortwave (Figures 5h, 5k, and 5n). Similarly to the qualitative assessment in Figure 4, scatterplots of top-of-atmosphere upwelling, and surface downwelling (Figures 5c, 5f, 5i, 5l, and 5o) flux predictions are close to reference calculations across the range of values. The third column in Figure 6 shows larger errors in the vertical structure of the 3D effects on heating rates (Figures 6c and 6f). The size of 3D effects on heating rates (Figures 6b and 6e) is, however, about two orders of magnitude smaller than the absolute heating rates from SPARTACUS (Figures 6a and 6d).

Table 2 summarizes bulk error statistics for fluxes and heating rates. NNs errors are generally small. The mean percentage error is below 20 %, except for the upwelling shortwave where it is −96 %. This latter result is not particularly interesting, however, as the mean 3D cloud effect for the entire upwelling component is about −0.16 W m$^{-2}$—much smaller than that at the top of the atmosphere of −1.3 W m$^{-2}$. The mean absolute percentage error of fluxes is about 20 %–30 %; in other words, NN predictions capture about 70 %–80 % of the 3D effects predicted by SPARTACUS. For heating rates, the mean and mean absolute percentage errors are 15 % and 66 % for the longwave, and −6.1 % and 62 % for the shortwave. This latter result is not particularly important as 3D effects on heating rates are small, about 0.01 K d$^{-1}$ for the shortwave. Indeed, the primary means by which shortwave 3D effects influence the Earth system is via a change in surface fluxes, and from there the surface temperature.

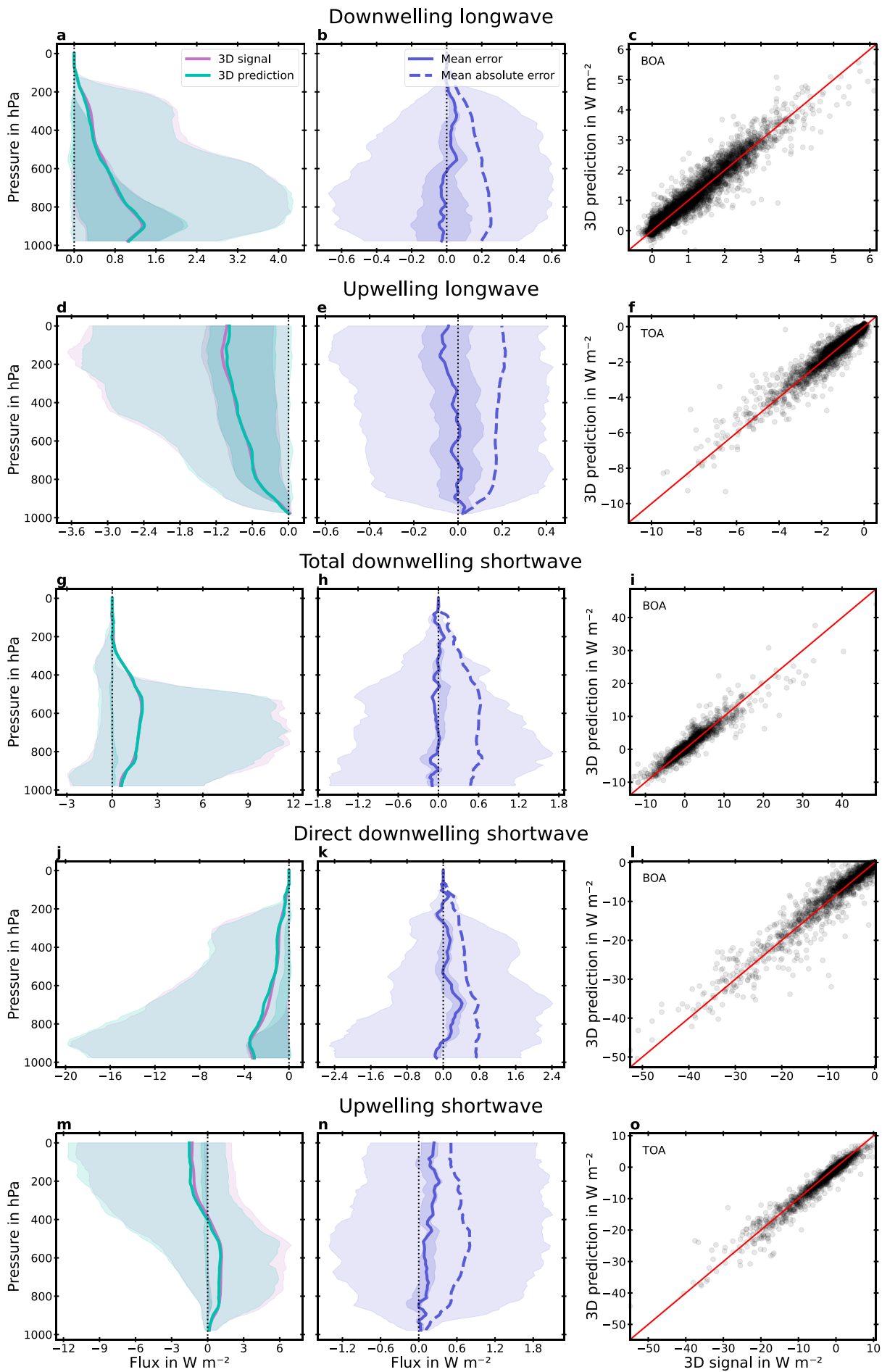

**Figure 5.** Per-level statistics of mean (**a,d,g,j,m**) 3D signal (SPARTACUS minus Tripleclouds) and 3D prediction (NN), (**b,e,h,k,n**) mean (continuous) and mean absolute (dashed) error (3D prediction minus 3D signal), and (**c,f,i,l,o**) scatterplots of 3D top-of-atmosphere (TOA) and bottom-of-atmosphere (BOA) fluxes, computed using the test fraction. 50 % (lighter) and 90 % (darker) quantiles are shown for right and middle panels.

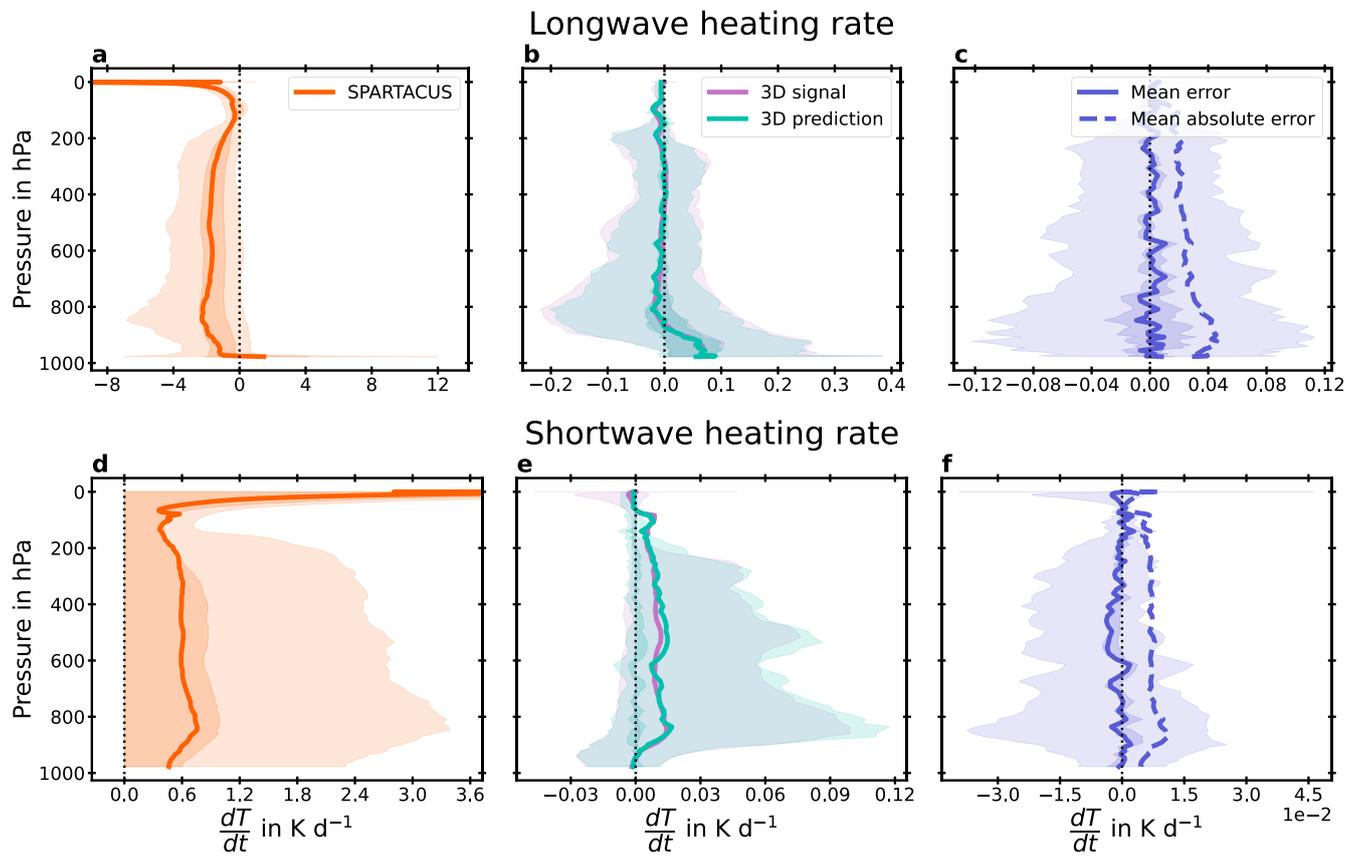

**Figure 6.** Per-level statistics of (**a**,**d**) mean absolute heating rates from SPARTACUS, (**b**,**e**) mean 3D signal and prediction, and (**c**,**f**,) mean (continuous) and mean absolute (dashed) error (3D prediction minus 3D signal), computed using the test fraction. 50 % (lighter) and 90 % (darker) quantiles are shown for all panels.

**Table 2.** Bulk mean and mean absolute 3D signal (SPARTACUS minus Tripleclouds), 3D error (prediction minus signal), and percentage error (3D error divided by 3D signal, multiplied by 100) for (**a**) longwave ($L$) and shortwave ($S$) fluxes, and (**b**) corresponding heating rates ($L^H$, $S^H$). Flux components are shown for the total upwelling (↑), downwelling (↓), and direct downwelling (⇓), and separately for top-of-atmosphere (TOA), and bottom-of-atmosphere (BOA).

|  | Mean | | | Mean Absolute | | |
|---|---|---|---|---|---|---|
|  | 3D signal | 3D error | Percentage error | 3D signal | 3D error | Percentage error |
| *(a) Fluxes* | | | | | | |
|  | W m$^{-2}$ | W m$^{-2}$ | % | W m$^{-2}$ | W m$^{-2}$ | % |
| $L^↓$ | 0.55 | 0.00048 | 0.087 | 0.56 | 0.14 | 25 |
| $L^↓_{BOA}$ | 1.0 | -0.029 | -2.8 | 1.1 | 0.2 | 19 |
| $L^↑$ | -0.72 | -0.025 | 3.4 | 0.73 | 0.17 | 23 |
| $L^↑_{TOA}$ | -1.1 | -0.071 | 6.7 | 1.1 | 0.2 | 19 |
| $S^↓$ | 0.73 | -0.032 | -4.4 | 1.1 | 0.38 | 33 |
| $S^↓_{BOA}$ | 0.53 | -0.092 | -17 | 1.4 | 0.48 | 35 |
| $S^⇓$ | -1.5 | 0.037 | -2.5 | 1.5 | 0.45 | 30 |
| $S^⇓_{BOA}$ | -3.2 | -0.15 | 4.5 | 3.2 | 0.72 | 22 |
| $S^↑$ | -0.16 | 0.15 | -96 | 1.6 | 0.52 | 32 |
| $S^↑_{TOA}$ | -1.3 | 0.22 | -17 | 1.9 | 0.51 | 27 |
| *(b) Heating Rates* | | | | | | |
|  | K d$^{-1}$ | K d$^{-1}$ | % | K d$^{-1}$ | K d$^{-1}$ | % |
| $L^H$ | 0.0069 | 0.001 | 15 | 0.037 | 0.024 | 66 |
| $S^H$ | 0.0066 | -0.0004 | -6.1 | 0.0099 | 0.0062 | 62 |

## 3.3 Runtime Performance

The emulators' runtime performance is assessed using a normalized runtime performance metric defined as total model runtime divided by total number of profiles. The total model runtime includes measurements of data normalization, inference with TensorFlow in Python, data denormalization, and postprocessing (Appendix A). To reduce input, output, and runtime overheads during measurements, the input data is replicated 10 times and the batch size of TensorFlow set to 50 000. Both ecRad and the NN are run fully single-threaded in Singularity (Kurtzer et al., 2017) with Ubuntu 18.04, GNU Fortran 7.5.0 compiler, Anaconda Python 3.8 and TensorFlow 2.4.1 on a shared AMD EPYC 7742 node with 32 CPUs and 124 GiB of system memory. SPARTACUS is about 4.58 times slower than Tripleclouds with an average of 11.6 ± 0.0196 ms per profile, compared to 2.53 ± 0.00854 ms for Tripleclouds. In comparison, the two NNs predicting 3D effects take 0.0257 ± 0.0000372 ms per profile. Thus, the combined time for running both Tripleclouds and the two NNs is 2.56 ± 0.00856 ms per profile, an increase of about 1.19 % of Tripleclouds' runtime. A key reason for the NNs being so fast is that they predict broadband quantities directly, rather than integrating over many spectral intervals (140 in the longwave and 112 in the shortwave) as done in Tripleclouds and SPARTACUS. While these absolute runtimes are expected to change when run on different hardware, or coupled to the IFS, relative differences are indicative of the order of speedup. With graphics processing units (GPUs) likely playing a significant role in future high-performance computing systems (Bauer et al., 2021), switching to GPUs is generally a trivial task with ML libraries such as TensorFlow.

## 4. Conclusion

In this paper we propose a hybrid physical machine learning approach to correct a fast but less accurate 1D radiative transfer scheme with two neural network emulators of shortwave and longwave 3D cloud effects. The emulators are trained on the difference between a 3D (SPARTACUS) and a 1D (Tripleclouds) solver. Results show that the 3D effects on fluxes are captured with bulk mean absolute errors between 20 % and 30 % of the 3D signal (Figures 3–5; Table 2). To put these results into perspective, Hogan et al. (2019) report the same error range, albeit with biases of about 0.3 W m$^{-2}$, for comparing the shortwave component of SPARTACUS to Monte Carlo simulations of 65 3D cloud scenes. Although profiles of heating rates show large mean absolute errors of up to 66 % (Table 2), the impact of 3D cloud effects on heating rates is up to two orders of magnitude smaller than that of the absolute heating rates (cf. Figures 6a–6d vs. Figures 6c–6f). As the 3D effects for top-of-atmosphere upwelling fluxes and surface downwelling fluxes are constantly improved, this hybrid physical machine learning approach may be valuable in operational settings where the computational performance of a parameterization scheme is often a limiting factor for its uptake. Here, clear-sky fluxes are efficiently and accurately computed using Tripleclouds, and cloudy profiles are corrected with neural network emulators that have a negligible impact on Tripleclouds' runtime performance (~1%; Section 3.3).

Although further improvements in emulating radiative transfer processes may be achieved with other types of network architectures (e.g., Ukkonen, 2021), the use of large domain-specific datasets such as those recently published as part of the MAchinE Learning for Scalable meTeoROlogy and climate project (see A3 in Dueben et al., 2021), or of data augmentation strategies (e.g., as implemented by Meyer, Nagler, et al., 2021) may help to further improve the accuracy and generalization of current emulators. As the number of vertical levels in the current emulator is fixed, retraining may be necessary if levels in the atmospheric model increase. However, we expect this to be a minor limitation as changes in operational components are often on a much longer time scale (i.e., a few years) than those needed to retrain and retest emulators. While we show that the emulation of 3D cloud effects is a promising area of research, it is only the first step toward operationalization. As new model capabilities may only be used operationally at the ECMWF if found to improve forecast skills, online evaluations within the ECMWF Integrated Forecast System, need to assess our findings in the broader context on skill scores and numerical stability: compensating errors in cloud-radiation interactions mean that changes in their representation may degrade forecast scores unless accompanied by other modifications (Haiden et al., 2018; Martin et al., 2010) and further influence a model's stability. Current research highlights challenges with NN emulators coupled to Earth system models, reporting degraded performance and unstable simulations under some circumstances (Brenowitz & Bretherton, 2019; Rasp et al., 2018). While our recent experience in emulating gravity wave drag (Chantry et al., 2021) and urban land surface (Meyer et al., 2022) schemes was positive, long coupled evaluations are required to better assess these type of models for operational use.

**Code and Data Availability**

Software, data, and tools are archived with a Singularity (Kurtzer et al., 2017) image deposited on Zenodo as described in the scientific reproducibility section of Meyer et al. (2020). Users wishing to access (and reproduce) the results can download the data archive at https://doi.org/10.5281/zenodo.5113055 (Meyer, 2021) and optionally run Singularity on their local or remote systems.

**Author Contributions**

Conceptualization: R.H., P.D.; Data curation: D.M.; Formal analysis: D.M.; Investigation: D.M.; Methodology: D.M., R.H., P.D.; Software: D.M.; Resources: D.M.; Validation: D.M., P.D; Visualization: D.M., S.M.; Writing – original draft preparation: D.M., R.H., P.D.; Writing – review & editing: D.M., R.H., P.D., S.M..

**Acknowledgments**

We thank Robert Pincus for the interesting discussion that inspired this research project and three anonymous reviewers for their useful feedback and contributions. Peter Dueben gratefully acknowledges funding from the Royal Society for his University Research Fellowship, as well as from the ESiWACE Horizon 2020 project (#823988) and the MAELSTROM EuroHPC Joint Undertaking project (#955513).

**Appendix A: Postprocessing Methods**

As introduced in Section 2.2, it can be challenging to use NNs to predict flux and heating rate profiles that are both, physically consistent with each other, and with heating rate profiles free from excessive noise. Here we describe a method to obtain consistent profiles by postprocessing NN outputs. Rather than using NNs to predict the profiles of 3D effects on upward and downward fluxes, we use them to predict the profiles of 3D effects on scalar fluxes (equal to the downwelling plus upwelling) and 3D effects on heating rates. As the latter are proportional to the divergence of the 3D effect on the net flux (downwelling *minus* upwelling), the information content is the same, but it is expressed in variables that are closer to what we need, and it is easier for the NNs to predict. For the rest of this appendix we omit the term "3D effect on" prefix for describing fluxes and heating rates. As the postprocessing method is common across longwave and shortwave components, we explain the main method via the longwave and highlight differences in assumptions and processing separately at the end of the section.

The starting point is the output from the neural network: the scalar flux profile at half levels $L^s = L^\downarrow + L^\uparrow$ (where $L^\downarrow$ and $L^\uparrow$ are the downwelling and upwelling fluxes) and the heating rate profile at full levels $H = -\frac{c_p}{g}\frac{\Delta L^n}{\Delta p}$, where $L^n = L^\downarrow - L^\uparrow$ is the net flux, $\Delta$ denotes the difference between the base and top of a layer so $\Delta p$ is the pressure difference across a layer, and $c_p$ and $g$ are the specific heat of dry air and the gravitational acceleration. The postprocessing consists of the following steps:

1. Compute the total atmospheric flux divergence (i.e., total emission minus absorption, in W m⁻²) from a heating rate profile. Fundamentally the divergence is the difference in net flux between the bottom-of-atmosphere (BOA) and top-of-atmosphere (TOA), i.e., $D = L^n_{\text{BOA}} - L^n_{\text{TOA}}$. To obtain this from the heating rate, we sum the profile of divergences of individual layers, that is, $D^H = \sum \Delta L^n$, where the $\Delta L^n$ profile is obtained from the heating rate by inverting the expression for $H$ above.

2. Compute the total atmospheric flux divergence from the scalar fluxes. At TOA, the downwelling longwave flux is zero so $L^n_{\text{TOA}} = -L^\uparrow = -L^s_{\text{TOA}}$ (in the shortwave the same formula can be applied because, even though the downwelling shortwave flux is not zero at TOA, the 3D effect on this part is). At BOA, the upwelling longwave flux is dominated by surface emission rather than reflection, so we can assume that the 3D effect is zero, leading to to $L^n_{\text{BOA}} = L^\downarrow = L^s_{\text{BOA}}$. Therefore, the atmospheric divergence estimated from the scalar fluxes is $D^s = L^s_{\text{BOA}} + L^s_{\text{TOA}}$.

3. Rescale the heating rate profile so that its divergence equals that from the scalar flux. This is done by multiplying the heating rates by a scaling factor equal to $D^s/D^H$, and, if necessary, capping the scaling factor to lie in the range 0.5 to 2. If capping, the scalar fluxes are also scaled to ensure that they have the same divergence.

4. Use the rescaled heating rate (and hence $\Delta L^n$) and scalar flux profiles to compute the profiles of upwelling and downwelling flux. First the $L^n$ profile is computed by integrating $\Delta L^n$ down from TOA from a start value of $L^n_{\text{TOA}} = -L^s_{\text{TOA}}$. Then the upwelling and downwelling components are computed from $L^\uparrow = \frac{L^s - L^n}{2}$ and $L^\downarrow = \frac{L^s + L^n}{2}$.

The calculation of shortwave components follows that of the longwave above, except for computing the BOA net flux from the scalar flux in step 2. The net shortwave flux is given by $S^n = S^\downarrow - S^\uparrow$, the scalar shortwave flux by $S^s = S^\downarrow + S^\uparrow$, and the albedo by $\alpha = S^\uparrow_{\text{BOA}}/S^\downarrow_{\text{BOA}}$, thus $S^n_{\text{BOA}} = S^s_{\text{BOA}}[(1-\alpha)/(1+\alpha)]$. As the total atmospheric flux divergence is the BOA net flux minus the TOA net flux, the total atmospheric flux divergence is computed from the shortwave scalar fluxes as $D^s = S^s_{\text{BOA}}[(1-\alpha)/(1+\alpha)] + S^s_{\text{TOA}}$. Other steps are identical to those for the longwave.